\pdfoutput=1

\documentclass[11pt]{article}

\usepackage[]{acl}

\usepackage{times}
\usepackage{latexsym}

\usepackage[T1]{fontenc}

\usepackage[utf8]{inputenc}

\usepackage{microtype}

\usepackage{graphicx}
\usepackage{multirow}
\usepackage{makecell}
\usepackage{hyperref}

\title{Addressing Issues of Cross-Linguality in Open-Retrieval Question Answering Systems For Emergent Domains}

\author{Alon Albalak, Sharon Levy, William Yang Wang \\ University of California, Santa Barbara \\ \texttt{\{alon\_albalak,sharonlevy,william\}@cs.ucsb.edu}}

\begin{document}
\maketitle
\begin{abstract}
Open-retrieval question answering systems are generally trained and tested on large datasets in well-established domains. However, low-resource settings such as new and emerging domains would especially benefit from reliable question answering systems. Furthermore, multilingual and cross-lingual resources in emergent domains are scarce, leading to few or no such systems.
In this paper, we demonstrate a cross-lingual open-retrieval question answering system for the emergent domain of COVID-19.
Our system adopts a corpus of scientific articles to ensure that retrieved documents are reliable. To address the scarcity of cross-lingual training data in emergent domains, we present a method utilizing automatic translation, alignment, and filtering to produce English-to-all datasets.
We show that a deep semantic retriever greatly benefits from training on our English-to-all data and significantly outperforms a BM25 baseline in the cross-lingual setting.
We illustrate the capabilities of our system with examples and release all code necessary to train and deploy such a system\footnote{Code provided at \href{https://github.com/alon-albalak/XOR-COVID}{github.com/alon-albalak/XOR-COVID}.}. 

\end{abstract}


\section{Introduction}

\begin{figure*}[t]
    \centering
    \includegraphics[width=0.95\textwidth]{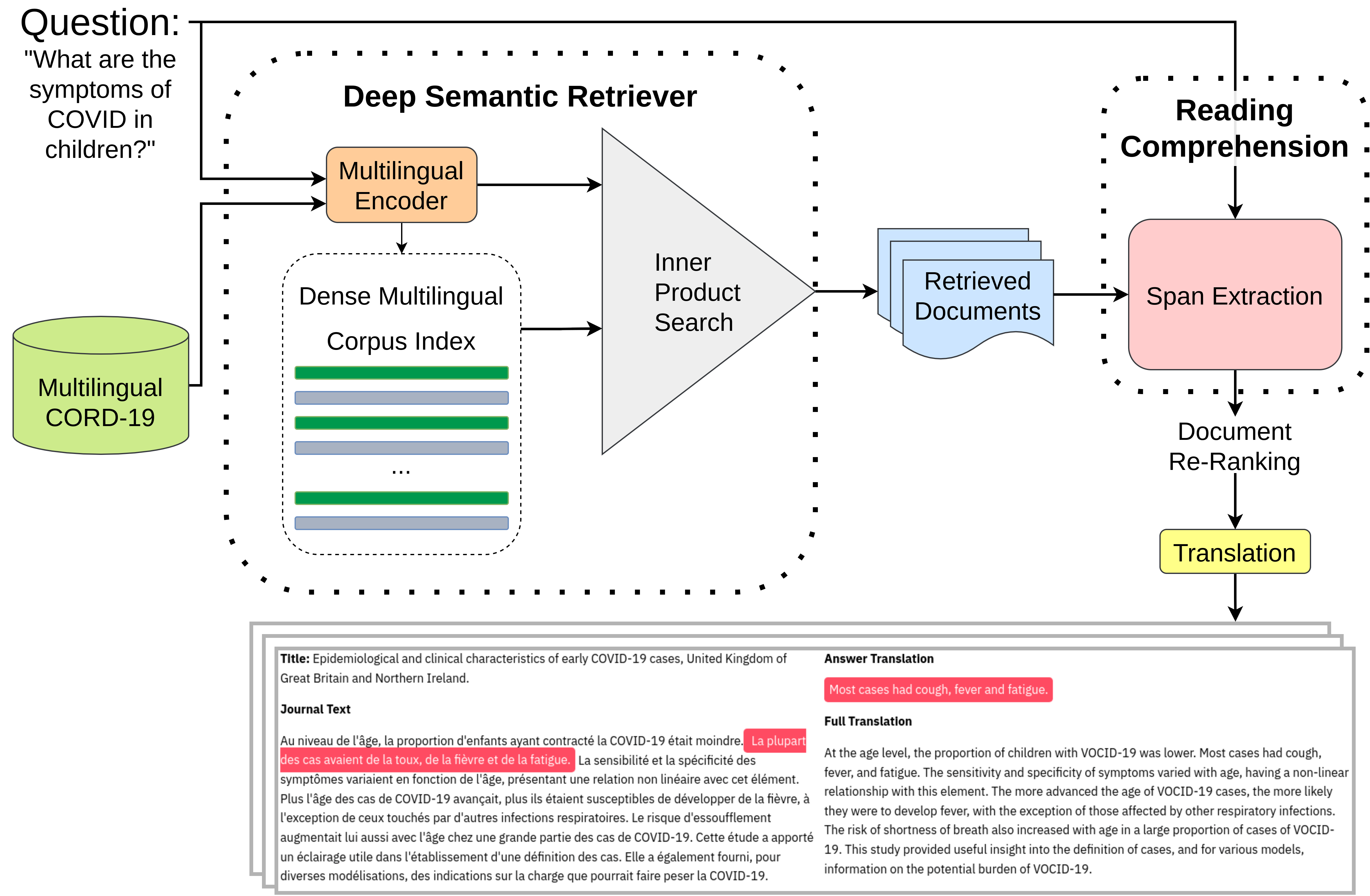}
    \caption{An overview of our cross-lingual COVID-19 open-retrieval question-answering system.}
    \label{fig:system_overview}
\end{figure*}

One challenge of emergent domains is that the originating locality is unknown, leading to the need for reliable information to cross language barriers. However, it is unlikely that domain-specific information will be available across multiple languages for a new domain.
Furthermore, information rapidly changes in emerging domains, compounding the challenge of accessing credible data.

An example of a prominent emergent domain is COVID-19, which has quickly spread across the globe. To combat the spread of misinformation about COVID-19, researchers have developed open-retrieval question answering \cite{chen-yih-2020-open} systems which use large collections of trusted documents. For example, \citet{lee2020answering}, \citet{levy-etal-2021-open}, and \citet{Esteva2021} all develop open-retrieval QA systems using large corpuses of scientific journal articles. However, because these systems focus on English, they leave a gap for implementation on emergent domains that do not originate in English-speaking locations.

To address the limitations of prior systems, we implement a cross-lingual open-retrieval question answering system that retrieves answers from a large collection of multilingual documents, where answers may be in a language different from the question \cite{xorqa}. 

In this work we take COVID-19 as an exemplar of an emergent domain and present our system, which addresses two main areas of importance:
\begin{itemize}
    \item \textit{Cross-linguality}: The locality of an emergent domain is unknown ahead of time, making cross-lingual QA essential. Additionally, because data can rapidly change in emerging domains, new information may develop in multiple languages, motivating the need for systems that work across many languages.
    \item \textit{Scarcity of training data}: Data scarcity is an expected concern for emergent domains, but multilingual and cross-lingual data are even more limited. We demonstrate that by employing automatic translation, alignment, and filtering methods, this challenge can be overcome in low-resource open-retrieval QA.
\end{itemize}

This system demonstration provides in-depth technical descriptions of the individual components of our cross-lingual open-retrieval question answering system: cross-lingual retrieval and cross-lingual reading comprehension modules. Then, we describe how to combine the components along with document re-ranking into the complete system, shown in Figure \ref{fig:system_overview}, and present several examples taken from our system.
\section{Cross-Lingual Dense Retrieval}
\label{sec:retrieval}


Training a dense retriever is challenging in low-resource settings, such as emergent domains, due to the data-hungry nature of large language models. This challenge is compounded 
in the cross-lingual setting, where we aim to train a model to encode concepts from multiple languages into a similar location in the embedding space. In this section, we discuss how we overcome these challenges.

\subsection{Data}
Cross-lingual retrieval requires two datasets; a large-scale multilingual corpus of scientific articles from which to retrieve documents and a cross-lingual dataset for training the retriever. However, a very limited number of COVID-19 datasets have been released, few of which are multilingual and none of which are cross-lingual.

\begin{figure}[t]
    \centering
    \includegraphics[width=\linewidth]{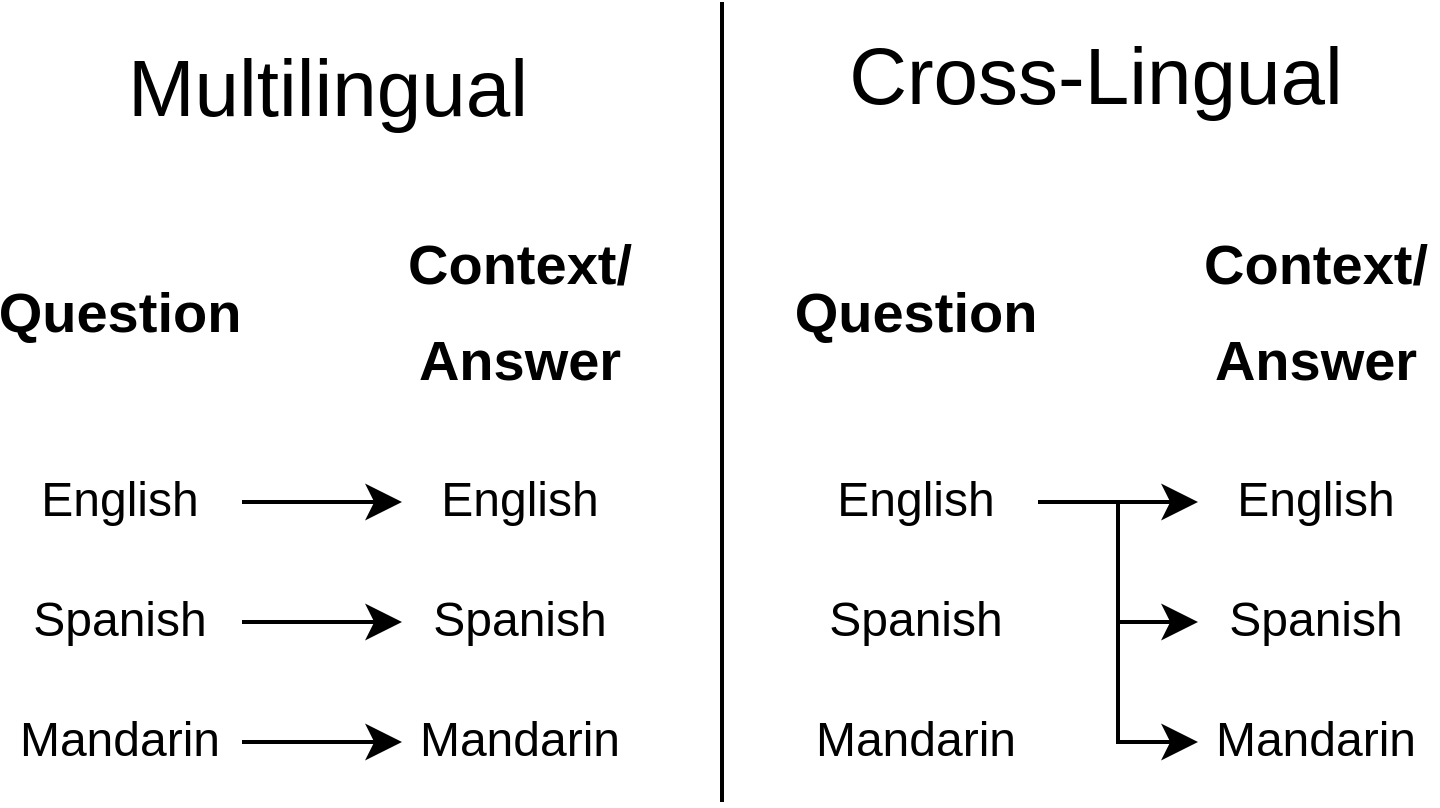}
    \caption{\textbf{Multilingual vs. cross-lingual question answering}: In the multilingual setting, QA pairs exist for multiple languages in a one-to-one mapping. On the other hand, in cross-lingual QA questions may have answers in any language, creating a one-to-many mapping.}
    \label{fig:multilingual_vs_crosslingual}
\end{figure}

\begin{table*}
\small
\centering
\begin{tabular}{|c|c|c|c|c|c|c|}
     \hline 
     \textbf{COUGH} & 9151 (en) & 1077 (es) & 778 (zh) & 697 (fr) & 573 (ja) & 531 (ar)  \\
     \hline
     \textbf{mCORD-19} & 172977 (en) & 1109 (es) & 951 (zh) & 711 (de) & 614 (fr) & 328 (pt) \\
     \hline
\end{tabular}
\caption{Top 6 languages by count for COUGH and the multilingual CORD-19 datasets. Language codes are the following: en-English, es-Spanish, zh-Chinese, fr-French, de-German, ja-Japanese, ar-Arabic, pt-Portuguese.}
\label{tab:lang_dist}
\end{table*}

\begin{table*}
\small
\centering
\begin{tabular}{|c|c|c|c|c|c|c|c|c|}
     \hline 
     \begin{tabular}{c}\textbf{Answer}\\\textbf{Language}\end{tabular} & \textbf{Spanish} & \textbf{Mandarin} & \textbf{French} & \textbf{Arabic} & \textbf{German} & \textbf{Russian} & \textbf{Vietnamese} & \textbf{Italian} \\
     \hline\hline
     \textbf{En2All} & 8695 & 8441 & 8372 & 8231 & 8226 & 8156 & 8072 & 8003 \\
     \hline
     \begin{tabular}{c}\textbf{Filtered}\\\textbf{En2All}\end{tabular} & 6620 & 5869 & 5635 & 5808 & 5867 & 4137 & 531 & 6568 \\
     \hline
\end{tabular}
\caption{QA pairs in our En2All and Filtered En2All variants of the COUGH dataset, where each question is in English, and the context and answer are in the language specified above.}
\label{tab:en2all_lang_dist}
\end{table*}

CORD-19 \cite{Wang2020CORD19TC} is a large-scale corpus of scientific papers on COVID-19, however a known limitation is that it contains only English articles. We draw inspiration from this work to address the lack of a large scale corpus of multilingual COVID-19 scientific articles. For our system, we use a manually collected corpus of English abstracts from PubMed, some of which have parallel abstracts in additional languages. The corpus is collected using the same query as described by \citet{Wang2020CORD19TC}
. We call this corpus multilingual CORD-19 (mCORD-19), and the language distribution can be found in Table \ref{tab:lang_dist}.

To train our retriever we utilize the COUGH \cite{zhang2021cough} dataset, which is a multilingual FAQ retrieval dataset and consists of COVID-19 QA pairs. Although COUGH is multilingual, containing samples in 9 different languages, COUGH does not contain any cross-lingual QA pairs. The language distribution is shown in Table \ref{tab:lang_dist}.

\subsection{Cross-lingual Data Generation}
To address the lack of cross-lingual data in COUGH we introduce a modification of the dataset which we call English-to-all (En2All), where we convert the dataset from the multilingual to cross-lingual setting, as demonstrated in Figure \ref{fig:multilingual_vs_crosslingual}. Because we are interested in a system which will find non-English answers to English questions, we create En2All through two translation processes. First, we translate the answer portion of every QA pair from COUGH into eight languages: Arabic, French, German, Italian, Mandarin, Russian, Spanish, and Vietnamese. Secondly, we translate the question portion of all QA pairs from any of the above languages into English\footnote{All translations are generated by the MarianNMT system \cite{mariannmt} through the Huggingface Transformers \cite{wolf-etal-2020-transformers} library.}.

As machine translation models do not perform perfectly, there may be instances within En2All that contain poor translations. To resolve this problem, we utilize LaBSE~\cite{feng2020language}, an existing BERT-based sentence embedding model that encodes 109 languages into a shared embedding space. The model is utilized to compare the alignment of translations across different languages. We take the following steps to filter out any poor translations in the data:
\begin{enumerate}
    \item We step through the current En2All and calculate similarity scores between translated answers and their original English answers. To do this, we have eight different comparisons for each translated English QA pair.
    \item Once the similarity scores have been calculated, we remove translations that do not meet a threshold and are classified as poor translations.
\end{enumerate}
After going through these steps, roughly one-third of the data samples from En2All are removed for poor translations.

\subsection{Methodology: Deep Semantic Retriever}
\label{sec:semantic_retriever}
Our retrieval model is based on the dense passage retriever from \citet{karpukhin-etal-2020-dense}. In contrast to their work, we train a unified encoder that encodes both query and corpus into a shared space. For the encoder, we train the multilingual BERT (mBERT) \cite{devlin-etal-2019-bert} and XLM-RoBERTa (XLM-R) \cite{DBLP:journals/corr/abs-1911-02116} models. Both models have been pre-trained using a tokenizer which shares a vocabulary for over 100 languages, allowing the models to encode all languages into a shared space. We train these models on the FAQ retrieval task by maximizing the inner product of correct QA pairs and minimizing the inner product of within-batch incorrect pairs.\\

\subsection{Cross-Lingual Retrieval Evaluation}
\label{sec:retrieval_eval}
To evaluate our models in the large-scale open-retrieval setting we utilize the questions from COUGH and En2All as our queries and the mCORD-19 dataset for our retrieval corpus. Because we have no ground truth labels for correct documents, and indeed there may be some unanswerable questions given this corpus, we measure model quality through a fuzzy matching metric, Fuzzy Match at top k documents (FM@k). FM@k utilizes the multilingual Sentence-BERT model from \cite{reimers-gurevych-2019-sentence}\footnote{We use the 'paraphrase-multilingual-mpnet-base-v2' variant}. Each of the top k retrieved documents is split into it's component sentences and embedded using the sentence-BERT model. Next, each sentence is compared with the ground truth answer by calculating the cosine similarity with the reference answer embedding from COUGH. If any of the cosine similarities for that documents sentences are above a threshold, the document is evaluated as a positive retrieval.

\begin{table}[]
    \centering
    \small
    \begin{tabular}{lcc}
         \hline
         \textbf{Model} & \begin{tabular}{c}\textbf{COUGH}\\(FM@5/100)\end{tabular} & \begin{tabular}{c}\textbf{COUGH}\\\textbf{+En2All}\\(FM@5/100)\end{tabular} \\
         \hline
         BM25 & \multicolumn{2}{c}{18.6/41.4} \\
         \hline
         mBERT\textsubscript{base} & 22.8/49.5 & 26.4/50.7 \\ \quad + En2All &28.0/54.9 & 27.7/51.7 \\
         \hline
         XLM-R\textsubscript{base} & 25.0/51.3 & 28.1/51.6 \\
         \quad + En2All & 30.1/55.4 & 28.4/52.2 \\
         \begin{tabular}{c} + Filtered-\\ En2All\end{tabular} & \textbf{32.9/56.7} & \textbf{30.9/53.4}
         \\
         \hline
         XLM-R\textsubscript{large} & 30.5/56.6 & 29.8/53.2 \\
         \quad + En2All & 32.1/56.4 & 29.6/52.9 \\
         \hline
    \end{tabular}
    \caption{\textbf{Retrieval evaluation results}. All models are trained on COUGH and additional training data is denoted by "+". The middle column takes queries from COUGH, the right column from COUGH and En2All. For both columns, the retrieval corpus is mCORD. FM@5 and FM@100 are the fuzzy matching techniques proposed to determine open-retrieval accuracy described in section \ref{sec:retrieval_eval}. Because BM25 is not cross-lingual, we translate it's queries into all languages in order to fairly compare against our cross-lingual models.}
    \label{tab:retrieval_eval}
\end{table}

The results for our models and a BM25 baseline are found in Table \ref{tab:retrieval_eval}. Since a multilingual BM25 cannot perform cross-lingual retrieval, in order to fairly compare against cross-lingual models, we translate all queries into every other language in the mCORD corpus and then perform BM25 retrieval.

BM25 drastically under performs compared to encoder models and demonstrates the need for a dense retrieval model. Although encoder models outperform BM25 when trained on multilingual data (COUGH), they are further improved by training on cross-lingual data (En2All). Additionally, after filtering low quality translations from En2All, we see further improvement in performance.
\section{Cross-Lingual Reading Comprehension}
\label{sec:reading_comprehension}

\subsection{Data}
\label{sec:reading_comprehension_data}
To train our cross-lingual reading comprehension model, we would ideally use a cross-lingual covid-specific question answering dataset. However, similarly to cross-lingual retrieval no such dataset exists so we augment existing datasets.

\citet{DBLP:journals/corr/abs-1910-11856} introduced XQuAD, a multilingual QA dataset composed of 240 paragraphs and 1190 QA pairs from SQuAD v1.1 which have been professionally translated into 10 languages. We utilize XQuAD as a pretraining dataset before performing any training on covid-specific datasets\footnote{We open-source our models pretrained on XQuAD at \href{https://huggingface.co/alon-albalak}{https://huggingface.co/alon-albalak}}.
\citet{moller-etal-2020-covid} introduce Covid-QA, a covid-specific QA dataset consisting of 2019 question-answer pairs, however, it contains english-only data. We modify Covid-QA with translations from MarianMT \cite{mariannmt} to generate two dataset variants based on the multilingual and cross-lingual settings shown in Figure \ref{fig:multilingual_vs_crosslingual}: Multilingual Covid-QA (MCQA) and English-to-all (En2All). MCQA is a multilingual version of Covid-QA, created by translating all QA pairs into 9 languages to match those from XQuAD: Arabic, German, Greek, Spanish, Hindi, Mandarin, Romanian, Russian, and Vietnamese.
En2All is our cross-lingual variation of Covid-QA, in a similar spirit to the cross-lingual variant of COUGH. Because Covid-QA is english-only, to generate En2All we translate all contexts/answers into the same 9 languages as MCQA.

\subsection{Methodology: Span Extraction}
Similar to our dense semantic retriever, we train mBERT and XLM-RoBERTa models for our reading comprehension task. We formulate reading comprehension as a span extraction task, where each model learns to find start and end tokens which represent the answer span in a document.

\begin{table}[t]
    \centering
    \small
    \begin{tabular}{ccc}
        \hline
         \textbf{Model} & \begin{tabular}{c} \textbf{MCQA}\\(EM/F1)\end{tabular} & \begin{tabular}{c} \textbf{MCQA+En2All}\\(EM/F1)\end{tabular} \\
         \hline
         mBERT\textsubscript{base} & 20.0/57.5 & 19.6/55.4 \\
         \quad + XQuAD & 21.2/57.7 & 20.5/55.6\\
         \quad + En2All & 19.3/56.1 & 19.2/55.8\\
         \hline
         XLM-R\textsubscript{base} & 25.1/60.0 & 24.4/58.9 \\
         \quad + XQuAD & 26.7/61.6 & 26.1/61.3 \\
         \quad + En2All & 24.0/58.8 & 23.9/58.3 \\
         \hline
         XLM-R\textsubscript{large} & 26.5/\textbf{62.7} & 26.4/\textbf{62.2} \\
         \quad + XQuAD & \textbf{29.1}/62.1 & \textbf{29.0}/61.7 \\
         \quad + En2All & 26.3/61.1 & 26.6/60.8
         \\
         \hline
    \end{tabular}
    \caption{\textbf{Reading comprehension evaluation results}. All models are trained on MCQA, and additional training data is denoted by "+". The left column shows evaluation on a multilingual dataset where questions/contexts are always in the same language. The right column additionally evaluates on a cross-lingual dataset where questions are in english and context paragraphs may be in any language.}
    \label{tab:reading_comprehension_result}
\end{table}

\begin{figure*}[t]
    \centering
    \fbox{\includegraphics[width=\textwidth]{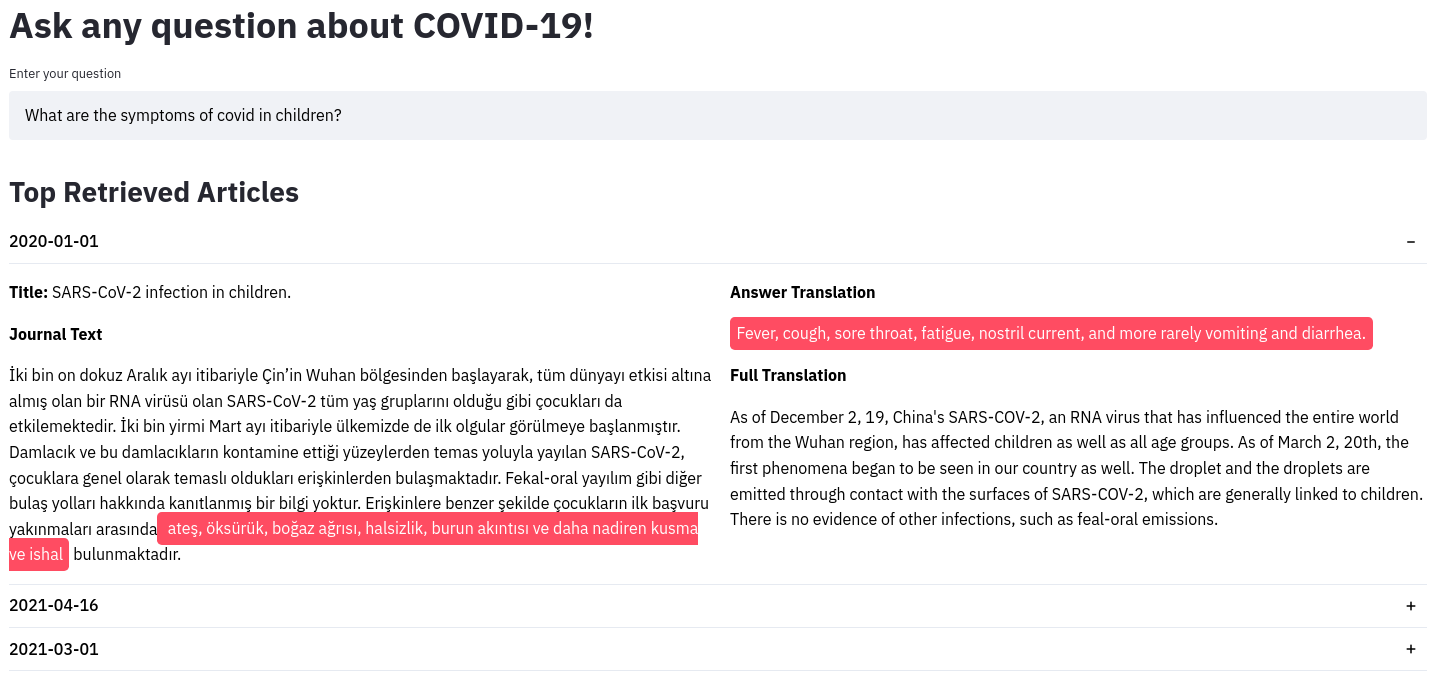}}
    \caption{\textbf{The main interface of our system}. At the top is the search bar, where the current query is "What are the symptoms of covid in children?" Below the search bar are the three retrieved articles, ranked by relevance. In this example, the first retrieved document has been expanded to show the title and original text in Turkish, on the left. And on the right is the translation of the answer and the full document into English.}
    \label{fig:sample_retrieved1}
\end{figure*}

\subsection{Cross-Lingual Reading Comprehension Evaluation}
To evaluate our models in the reading comprehension task, we utilize the QA datasets described in Section \ref{sec:reading_comprehension_data}. We evaluate our models based on exact match (EM) and F1 metrics by comparing the predicted answer spans with ground-truth answers.

The results for our models are found in Table \ref{tab:reading_comprehension_result}. We train each of our models on MCQA and supplement it with data from XQuAD or En2All. Interestingly, we find that although En2All improved models in the retrieval setting, it only hurt model performance in QA. We also see that pretraining on XQuAD improves performance in all metrics for both base models, but leads to a slight decrease in F1 score for XLM-R\textsubscript{large}. In our demo, we utilize XLM-R\textsubscript{large} which was pretrained on XQuAD because it has only slightly worse F1 score, but significantly higher exact match compared to the next best model.
\section{Cross-Lingual Open-Retrieval Question Answering}

\begin{figure*}
\centering
    \fbox{\includegraphics[width=\textwidth]{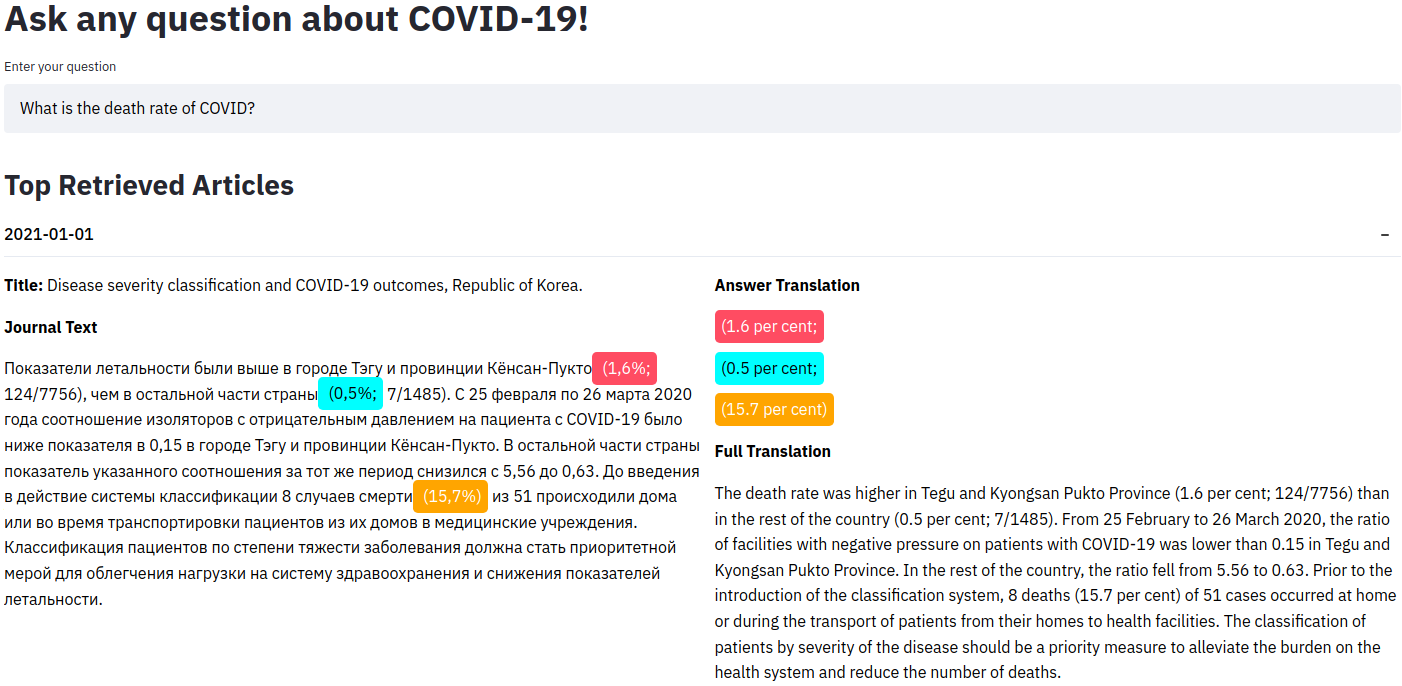}}
    \caption{A retrieved document for the query "What is the death rate of COVID", which shows multiple correct answers corresponding to different provinces of South Korea.}
    \label{fig:multiple_answer}
\end{figure*}

Our system is composed of the retrieval and reading comprehension modules described in sections \ref{sec:retrieval} and \ref{sec:reading_comprehension}. The full end-to-end system is shown in Figure \ref{fig:system_overview}. After the retriever has been trained, the mCORD-19 corpus is encoded and stored in the dense multilingual corpus index. When a question is posed to the system, the query is encoded, and a maximum inner product search is performed over the index to find documents most similar to the query. Answers are then extracted from the retrieved documents and the documents are re-ranked based on answer confidence from the span extraction model. Finally, the answer spans and full documents are translated into English and presented to the user with highlighted answers.
\section{Demo}
The demonstration retrieves documents from our mCORD-19 corpus, which has been encoded by the deep semantic retriever from section \ref{sec:semantic_retriever}. We provide additional examples from the demo in Appendix \ref{sec:appendix}.

\begin{figure}[t]
\centering
\includegraphics[width=0.55\linewidth]{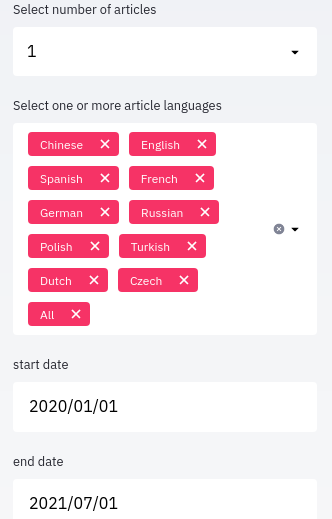}
\caption{The options sidebar for our demonstration system. The options include: number of articles to return, article languages to retrieve from, and publication date range. For visualization purposes we show all language options.
}
\label{fig:sidebar}
\end{figure}

\subsection{Sidebar Interface}
Our system has an options sidebar, shown in Figure \ref{fig:sidebar}, which gives the user several choices before entering a query. The user can determine how many documents they would like to see results from, they can select which languages the retrieved documents should be in, and they can specify a date range for the publications to search over. If there are no relevant documents in the desired date range, then the system will retrieve from any date range and displays a message to inform the user.

\subsection{Main Interface}
To query the system, a user simply selects the desired options from the sidebar and enters their question into the search bar, as seen in Figure \ref{fig:sample_retrieved1}. After the user enters their question, the system will encode the question using the trained deep semantic retriever and find the most relevant documents within the given language and date range constraints. Then, the reading comprehension model will extract the answer (or answers) most relevant to the query from each retrieved document. Additionally, for any non-English documents, the system translates both the retrieved article and extracted answers into English\footnote{All translations are generated by MarianNMT \cite{mariannmt} from the Huggingface Transformers library \cite{wolf-etal-2020-transformers}.}. Finally, the retrieved documents will be re-ranked based on the confidence scores for the extracted answers.

The desired number of documents will be displayed to the user as a list of publication dates. Each item can be expanded to show the article title, original document with highlighted answers, translated answers, and the full article translation. If an article contains a single answer, it will be highlighted in red. If there are multiple answers, each answer will be highlighted with a different color to allow for easy alignment between original answers and their translations, demonstrated in Figure \ref{fig:multiple_answer}. Additional examples are provided in Appendix \ref{sec:appendix}.
\section{Conclusion}
In this work, we tackled two challenging areas in open-retrieval QA: cross-linguality and data scarcity. We presented methods for generating cross-lingual data in an emergent domain, COVID-19. Then, we demonstrated that an open-retrieval QA system trained on our data significantly outperforms a BM25 baseline. We hope that the methods presented here allow for increased access to reliable information in future emergent domains.
\section{Ethics Statement}
Crucial to any open-retrieval question-answering system, the credibility and truthfulness of the documents is paramount, in particular when trying to prevent and combat misinformation that arises in emergent domains. Any question-answering system is limited by the corpus used. To this end, we do our best to ensure that any information included in our corpus is truthful by including only peer-reviewed scientific articles from PubMed\footnote{https://pubmed.ncbi.nlm.nih.gov/}.

Furthermore, there may be emergent domains without peer-reviewed scientific articles from which to draw answers. In these cases (and in fact in cases where peer-review does exist) it is imperative to include sources along with answers. This allows for users to judge the quality of information. In our system we present the title and date of publication for each returned article so that users can find the source content if desired.

\bibliography{anthology,custom}

\begin{thebibliography}{16}
\expandafter\ifx\csname natexlab\endcsname\relax\def\natexlab#1{#1}\fi

\bibitem[{Artetxe et~al.(2020)Artetxe, Ruder, and
  Yogatama}]{DBLP:journals/corr/abs-1910-11856}
Mikel Artetxe, Sebastian Ruder, and Dani Yogatama. 2020.
\newblock On the cross-lingual transferability of monolingual representations.
\newblock In \emph{ACL}.

\bibitem[{Asai et~al.(2021)Asai, Kasai, Clark, Lee, Choi, and
  Hajishirzi}]{xorqa}
Akari Asai, Jungo Kasai, Jonathan~H. Clark, Kenton Lee, Eunsol Choi, and
  Hannaneh Hajishirzi. 2021.
\newblock {XOR} {QA}: Cross-lingual open-retrieval question answering.
\newblock In \emph{NAACL-HLT}.

\bibitem[{Chen and Yih(2020)}]{chen-yih-2020-open}
Danqi Chen and Wen-tau Yih. 2020.
\newblock \href {https://doi.org/10.18653/v1/2020.acl-tutorials.8} {Open-domain
  question answering}.
\newblock In \emph{Proceedings of the 58th Annual Meeting of the Association
  for Computational Linguistics: Tutorial Abstracts}, pages 34--37, Online.
  Association for Computational Linguistics.

\bibitem[{Conneau et~al.(2020)Conneau, Khandelwal, Goyal, Chaudhary, Wenzek,
  Guzm{\'a}n, Grave, Ott, Zettlemoyer, and
  Stoyanov}]{DBLP:journals/corr/abs-1911-02116}
Alexis Conneau, Kartikay Khandelwal, Naman Goyal, Vishrav Chaudhary, Guillaume
  Wenzek, Francisco Guzm{\'a}n, Edouard Grave, Myle Ott, Luke Zettlemoyer, and
  Veselin Stoyanov. 2020.
\newblock \href {https://doi.org/10.18653/v1/2020.acl-main.747} {Unsupervised
  cross-lingual representation learning at scale}.
\newblock In \emph{Proceedings of the 58th Annual Meeting of the Association
  for Computational Linguistics}, pages 8440--8451, Online. Association for
  Computational Linguistics.

\bibitem[{Devlin et~al.(2019)Devlin, Chang, Lee, and
  Toutanova}]{devlin-etal-2019-bert}
Jacob Devlin, Ming-Wei Chang, Kenton Lee, and Kristina Toutanova. 2019.
\newblock \href {https://doi.org/10.18653/v1/N19-1423} {{BERT}: Pre-training of
  deep bidirectional transformers for language understanding}.
\newblock In \emph{Proceedings of the 2019 Conference of the North {A}merican
  Chapter of the Association for Computational Linguistics: Human Language
  Technologies, Volume 1 (Long and Short Papers)}, pages 4171--4186,
  Minneapolis, Minnesota. Association for Computational Linguistics.

\bibitem[{Esteva et~al.(2021)Esteva, Kale, Paulus, Hashimoto, Yin, Radev, and
  Socher}]{Esteva2021}
Andre Esteva, Anuprit Kale, Romain Paulus, Kazuma Hashimoto, Wenpeng Yin,
  Dragomir Radev, and Richard Socher. 2021.
\newblock \href {https://doi.org/10.1038/s41746-021-00437-0} {Covid-19
  information retrieval with deep-learning based semantic search, question
  answering, and abstractive summarization}.
\newblock \emph{npj Digital Medicine}, 4(1):68.

\bibitem[{Feng et~al.(2020)Feng, Yang, Cer, Arivazhagan, and
  Wang}]{feng2020language}
Fangxiaoyu Feng, Yinfei Yang, Daniel Cer, Naveen Arivazhagan, and Wei Wang.
  2020.
\newblock Language-agnostic bert sentence embedding.
\newblock \emph{arXiv preprint arXiv:2007.01852}.

\bibitem[{Junczys-Dowmunt et~al.(2018)Junczys-Dowmunt, Grundkiewicz, Dwojak,
  Hoang, Heafield, Neckermann, Seide, Germann, Fikri~Aji, Bogoychev, Martins,
  and Birch}]{mariannmt}
Marcin Junczys-Dowmunt, Roman Grundkiewicz, Tomasz Dwojak, Hieu Hoang, Kenneth
  Heafield, Tom Neckermann, Frank Seide, Ulrich Germann, Alham Fikri~Aji,
  Nikolay Bogoychev, Andr\'{e} F.~T. Martins, and Alexandra Birch. 2018.
\newblock \href {http://www.aclweb.org/anthology/P18-4020} {Marian: Fast neural
  machine translation in {C++}}.
\newblock In \emph{Proceedings of ACL 2018, System Demonstrations}, pages
  116--121, Melbourne, Australia. Association for Computational Linguistics.

\bibitem[{Karpukhin et~al.(2020)Karpukhin, Oguz, Min, Lewis, Wu, Edunov, Chen,
  and Yih}]{karpukhin-etal-2020-dense}
Vladimir Karpukhin, Barlas Oguz, Sewon Min, Patrick Lewis, Ledell Wu, Sergey
  Edunov, Danqi Chen, and Wen-tau Yih. 2020.
\newblock \href {https://doi.org/10.18653/v1/2020.emnlp-main.550} {Dense
  passage retrieval for open-domain question answering}.
\newblock In \emph{Proceedings of the 2020 Conference on Empirical Methods in
  Natural Language Processing (EMNLP)}, pages 6769--6781, Online. Association
  for Computational Linguistics.

\bibitem[{Lee et~al.(2020)Lee, Yi, Jeong, Sung, Yoon, Choi, Ko, and
  Kang}]{lee2020answering}
Jinhyuk Lee, Sean~S. Yi, Minbyul Jeong, Mujeen Sung, WonJin Yoon, Yonghwa Choi,
  Miyoung Ko, and Jaewoo Kang. 2020.
\newblock \href {https://doi.org/10.18653/v1/2020.nlpcovid19-2.1} {Answering
  questions on {COVID}-19 in real-time}.
\newblock In \emph{Proceedings of the 1st Workshop on {NLP} for {COVID}-19
  (Part 2) at {EMNLP} 2020}, Online. Association for Computational Linguistics.

\bibitem[{Levy et~al.(2021)Levy, Mo, Xiong, and Wang}]{levy-etal-2021-open}
Sharon Levy, Kevin Mo, Wenhan Xiong, and William~Yang Wang. 2021.
\newblock \href {https://doi.org/10.18653/v1/2021.emnlp-demo.30} {Open-{D}omain
  question-{A}nswering for {COVID}-19 and other emergent domains}.
\newblock In \emph{Proceedings of the 2021 Conference on Empirical Methods in
  Natural Language Processing: System Demonstrations}, pages 259--266, Online
  and Punta Cana, Dominican Republic. Association for Computational
  Linguistics.

\bibitem[{Lu~Wang et~al.(2020)Lu~Wang, Lo, Chandrasekhar, Reas, Yang, Eide,
  Funk, Kinney, Liu, Merrill, Mooney, Murdick, Rishi, Sheehan, Shen, Stilson,
  Wade, Wang, Wilhelm, Xie, Raymond, Weld, Etzioni, and
  Kohlmeier}]{Wang2020CORD19TC}
Lucy Lu~Wang, Kyle Lo, Yoganand Chandrasekhar, Russell Reas, Jiangjiang Yang,
  Darrin Eide, Kathryn Funk, Rodney Kinney, Ziyang Liu, William Merrill, Paul
  Mooney, Dewey Murdick, Devvret Rishi, Jerry Sheehan, Zhihong Shen, Brandon
  Stilson, Alex~D. Wade, Kuansan Wang, Chris Wilhelm, Boya Xie, Douglas
  Raymond, Daniel~S. Weld, Oren Etzioni, and Sebastian Kohlmeier. 2020.
\newblock \href {https://pubmed.ncbi.nlm.nih.gov/32510522} {Cord-19: The
  covid-19 open research dataset}.
\newblock \emph{ArXiv}, page arXiv:2004.10706v2.
\newblock 32510522[pmid].

\bibitem[{M{\"o}ller et~al.(2020)M{\"o}ller, Reina, Jayakumar, and
  Pietsch}]{moller-etal-2020-covid}
Timo M{\"o}ller, Anthony Reina, Raghavan Jayakumar, and Malte Pietsch. 2020.
\newblock \href {https://aclanthology.org/2020.nlpcovid19-acl.18} {{COVID-QA}:
  A question answering dataset for {COVID}-19}.
\newblock In \emph{Proceedings of the 1st Workshop on {NLP} for {COVID-19} at
  {ACL} 2020}, Online. Association for Computational Linguistics.

\bibitem[{Reimers and Gurevych(2019)}]{reimers-gurevych-2019-sentence}
Nils Reimers and Iryna Gurevych. 2019.
\newblock \href {https://doi.org/10.18653/v1/D19-1410} {Sentence-{BERT}:
  Sentence embeddings using {S}iamese {BERT}-networks}.
\newblock In \emph{Proceedings of the 2019 Conference on Empirical Methods in
  Natural Language Processing and the 9th International Joint Conference on
  Natural Language Processing (EMNLP-IJCNLP)}, pages 3982--3992, Hong Kong,
  China. Association for Computational Linguistics.

\bibitem[{Wolf et~al.(2020)Wolf, Debut, Sanh, Chaumond, Delangue, Moi, Cistac,
  Rault, Louf, Funtowicz, Davison, Shleifer, von Platen, Ma, Jernite, Plu, Xu,
  Scao, Gugger, Drame, Lhoest, and Rush}]{wolf-etal-2020-transformers}
Thomas Wolf, Lysandre Debut, Victor Sanh, Julien Chaumond, Clement Delangue,
  Anthony Moi, Pierric Cistac, Tim Rault, Rémi Louf, Morgan Funtowicz, Joe
  Davison, Sam Shleifer, Patrick von Platen, Clara Ma, Yacine Jernite, Julien
  Plu, Canwen Xu, Teven~Le Scao, Sylvain Gugger, Mariama Drame, Quentin Lhoest,
  and Alexander~M. Rush. 2020.
\newblock \href {https://www.aclweb.org/anthology/2020.emnlp-demos.6}
  {Transformers: State-of-the-art natural language processing}.
\newblock In \emph{Proceedings of the 2020 Conference on Empirical Methods in
  Natural Language Processing: System Demonstrations}, pages 38--45, Online.
  Association for Computational Linguistics.

\bibitem[{Zhang et~al.(2021)Zhang, Sun, Yue, Lin, and Sun}]{zhang2021cough}
Xinliang~Frederick Zhang, Heming Sun, Xiang Yue, Simon Lin, and Huan Sun. 2021.
\newblock {COUGH}: A challenge dataset and models for {COVID}-19 {FAQ}
  retrieval.
\newblock In \emph{Proceedings of the 2021 Conference on Empirical Methods in
  Natural Language Processing, {EMNLP} 2021}, pages 3759--3769.

\end{thebibliography}
\bibliographystyle{acl_natbib}

\appendix

\section{Additional Examples}
\label{sec:appendix}
\begin{figure*}
    \centering
    \fbox{\includegraphics[width=\textwidth]{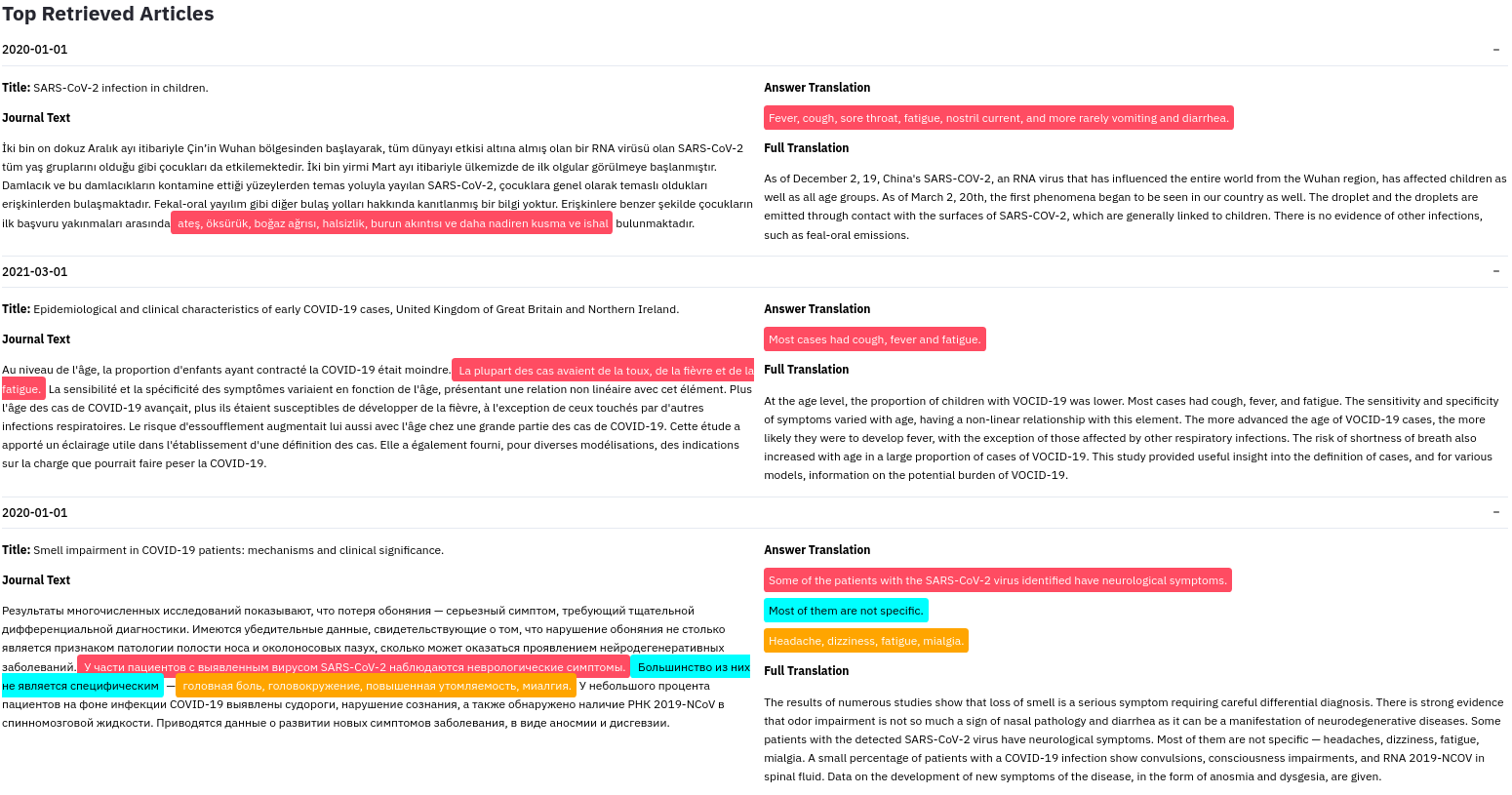}}
    \caption{The top 3 non-English results for the query "What are the symptoms of covid in children?"}
    \label{fig:additional_example1}
\end{figure*}

\begin{figure*}
    \centering
    \fbox{\includegraphics[width=\textwidth]{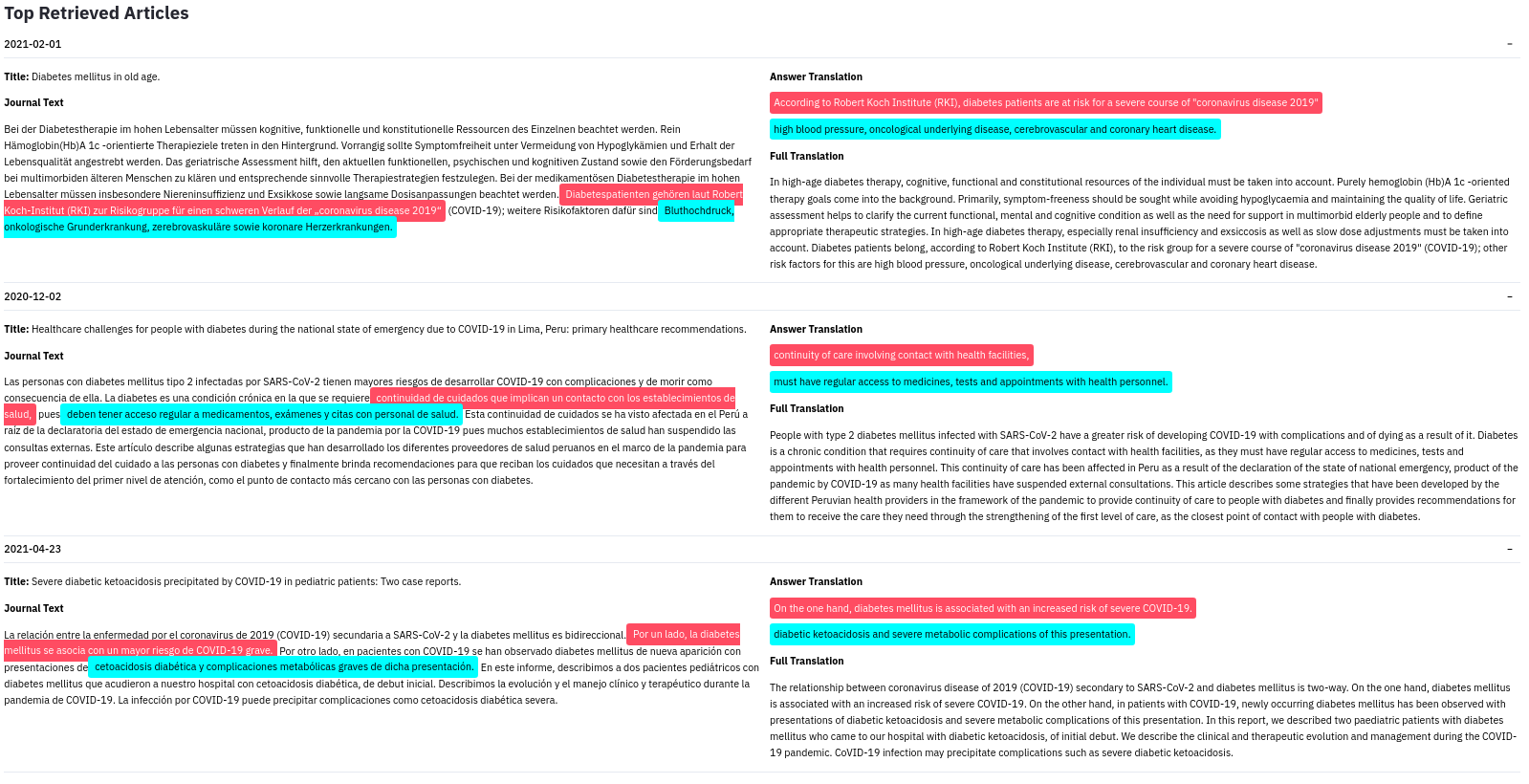}}
    \caption{The top 3 non-english results for the query "What are the concerns of having covid and diabetes?"}
    \label{fig:additional_example2}
\end{figure*}

\begin{figure*}
    \centering
    \fbox{\includegraphics[width=\textwidth]{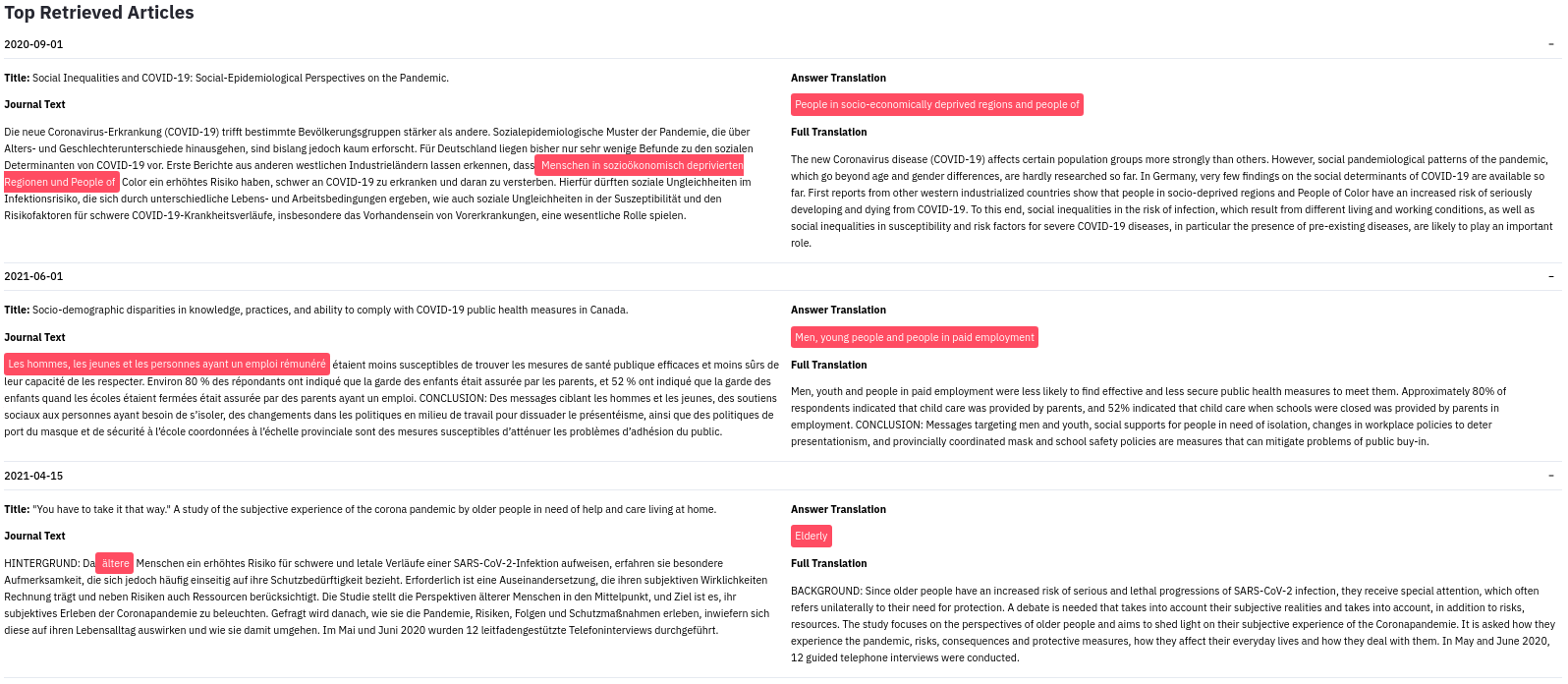}}
    \caption{The top-3 non-english results for the query "Who is most vulnerable to covid?"}
    \label{fig:additional_example3}
\end{figure*}

\end{document}